\documentclass{mep}
 \pdfoutput=1
\usepackage[utf8]{inputenc}
\usepackage[T1]{fontenc}
\usepackage{booktabs} 
\usepackage{graphicx}
\usepackage{hyperref}

\begin{document}

\title{Large e-retailer image dataset for visual search and product classification}

\addauthor{Arnaud Bellétoile for the Cdiscount datascience team}{datascience@cdiscount.com}{1}
\addinstitution{Cdiscount, France}

\maketitle

\begin{abstract}
Recent results of deep convolutional networks in visual recognition challenges open the path to a whole new set of disruptive user experiences such as visual search or recommendation. The list of companies offering this type of service is growing everyday but the adoption rate and the relevancy of results may vary a lot.

We believe that the availability of large and diverse datasets is a necessary condition to improve the relevancy of such recommendation systems and facilitate their adoption. For that purpose, we wish to share with the community this dataset of more than 12M images of the 7M products of our online store classified into 5K categories. 

This original dataset is introduced in this article and several features are described. We also present some aspects of the winning solutions of our image classification challenge that was organized on the Kaggle platform around this set of images.
\end{abstract}

\section{Introduction}

Recent advances in artificial intelligence and image recognition allow a whole new set of services to improve the Internet shopping experience~\cite{he2016deep, simonyan2015very}. Among those new services, visual search is probably one of the most promising technique as it provides an effective and natural way to search through a catalog with a simple picture~\cite{huang2015systems}.

Improving visual recommendation algorithms requires access to large labeled image datasets, possibly specialized in the core business they address. Available generic image datasets include TinyImage~\cite{Torralba08}, LabelMe~\cite{Russell08}, Lotus Hill~\cite{Yao07}, Microsoft Common Objects in Context~\cite{Lin14} (COCO) or OpenImages~\cite{Krasin17}. Of course, ImageNet~\cite{Jia09} is the \textit{de facto} standard to bench image classification algorithms involving extremely large numbers of labels.

Most of the public datasets that are of direct use to on-line retailers are specialized in fashion items: the Exact Street2Shop~\cite{Kiapour15} dataset identifies around 40,000 clothing items worn by people on real-world street photos, and provides their exact match amongst hundreds of thousands of images from shopping websites; DeepFashion consists of over 800,000 annotated images that contain clothes~\cite{Liu16}. 

To the best of our knowledge, no comprehensive image dataset covering products typically sold by generalist retailers is yet available to the community. This is the reason why we are releasing a large dataset of such categorized product images. With more than 12M images of 7M products classified into 5270 categories, this dataset should help the community to leverage state-of-the-art neural network architectures in order to develop better recommendation systems.

In the following article, we present several aspects of the dataset such as the way it was build and organized or some specific features that you might want to consider before training a model on it. We also get a grasp of the approaches followed by the 3 winning teams of the Kaggle competition we organized on this dataset.

The full dataset can be downloaded from the Cdiscount challenge page on the Kaggle platform~\cite{kaggle} (the url is given in the reference).

\section{E-retailer product catalog}\label{sec-dataDesc}

The large e-retailer image dataset we present has been extracted from the full list of products available on our web site in July, 2017. Therefore, products may be coming from our own list of products or from our Market Place where independent resellers can put their products up for sell. Our own catalog being rich of approximately 200,000~products, the vast majority of the 7M products in the dataset originates from the nearly 10,000 independent resellers present on our Market Place.

Our catalog is organized according to a 3-level aggregation tree with French labeled categories. The $\mathrm{1^{st}}$ level of aggregation is referred to as \textit{Cat~I} for Category level I and contains a diversity of products that could be compared to a physical store like a drugstore or a wine shop. It is the most generic level of aggregation and as such is of particular interest if one wishes to focus on a particular subset of images such as \textit{CHILDCARE~(PUERICULTURE)}, \textit{BAGS~(BAGAGERIE)} or \textit{INTERIOR DESIGN~(DECORATION)}. The 49 distinct Cat~III in the data set are listed in the table~\ref{tab:store} with the corresponding English translation.

\begin{table}[!htbp]

  \begin{tabular}{l|r}
\textbf{Cat I} & \textbf{Translation} \\
\hline \hline
ABONNEMENT / SERVICES & SUBS. / SERVICES \\
AMNG URB.-VOIRIE & URBAN PLANING-ROADWAY \\
ANIMALERIE & PET SHOP \\
APICULTURE & APICULTURE \\
ART DE LA TABLE-ART. CUL. & TABLEWARE-COOK. UST. \\
ARTICLES POUR FUMEUR & SMOKER TOOLS \\
AUTO-MOTO & CAR-MOTORCYCLE \\
BAGAGERIE & BAGS \\
BATEAU MOTEUR-VOILIER & MOTOR BOAT-SAILING  BOAT \\
BJX-LUNETTES-MONTRES & JEWEL.-GLASS.-WATCHES \\
BRICO.-OUTIL.-QUINC. & DIY-TOOLING-HARDWARE\\
CHAUSSURES-ACCESS. & SHOES-ACCESSORIES \\
COFFRET CADEAU BOX & GIFT BOX \\
CONDITIONNEMENT & PACKAGING \\
DECO-LINGE-LUMINAIRE & INT. DESIGN-TEXT.-LUM \\
DROGUERIE & DRUGSTORE \\
DVD-BLURAY & DVD-BLURAY \\
ELECTROMENAGER & APPLIANCES \\
ELECTRONIQUE & ELECTRONIC \\
EPICERIE & GROCERY \\
FUNERAIRE & FUNERAL \\
HYGIENE-BEAUTE-PARFUM & HYG-BEAUTY-PERF \\
INFORMATIQUE & COMPUTER EQUIPMENT \\
INSTRUMENTS DE MUSIQUE & MUSICAL INSTRUMENT \\
JARDIN-PISCINE & GARDEN-SWIMMING POOL \\
JEUX-JOUETS & GAMES-TOYS \\
JEUX VIDEO & VIDEO GAMES \\
LIBRAIRIE & BOOKSTORE \\
LITERIE & BEDDING \\
LSRS CREA.-BX ARTS-PAPET. & CRAFT-ARTS-STATIONERY \\
MANUTENTION	MATERIAL & HANDLING \\
MATERIEL DE BUREAU & OFFICE EQUIPMENT \\
MATERIEL MEDICAL & MEDICAL DEVICE \\
MERCERIE & HABERDASHERY \\
MEUBLE & FURNITURE \\
MUSIQUE & MUSIC \\
PARAPHARMACIE & DRUGSTORE \\
PHOTO-OPTIQUE & PHOTO-OPTIC \\
PT DE VTE-COMM.-ADMIN. & SALES OUT-COMM.-ADMIN. \\
PRODUITS FRAIS & FRESH PRODUCES \\
PRODUITS SURGELES & FROZEN FOODS \\
PUERICULTURE & CHILDCARE \\  
SONO-DJ & SOUND SYSTEM-DJ \\
SPORT & SPORT \\
TATOUAGE-PIERCING & TATOO-PIERCING \\
TELEPHONIE-GPS & TELEPHONY-GPS \\
TENUE PROFESSIONNELLE & WORKING CLOTHES \\
TV-VIDEO-SON & TV-VIDEO-HIFI \\
VIN-ALCOOL-LIQUIDES	& WINE-ALCOHOL-LIQUIDS
  \end{tabular}
  \caption{French labeled categories of the first level of aggregation in the dataset and the corresponding translations.}
  \label{tab:store}
\end{table}

\begin{table}[!htbp]
\center
	\begin{tabular}{l||ccc}
    \textbf{Category level} & Cat~I & Cat~II & Cat~III \\
    \hline
    \textbf{Nb of categories} & 49 & 483 & 5270 \\
    \end{tabular}
    \caption{Number of values taken by the 3 levels of the categorization tree.}
    \label{table:cat}
\end{table}

\begin{table}[!htbp]
\center
	\begin{tabular}{l||cccc}
    \textbf{Nb of images} & \#1 & \#2 & \#3 & \#4\\
    \hline
    \textbf{Nb of products} & 4,369,441 & 1,128,588 & 542,792 & 1,029,075 \\
    \end{tabular}
    \caption{Number of products having exactly 1, 2, 3 or 4 associated images.}
    \label{table:img}
\end{table}

The $\mathrm{2^{nd}}$ level category~(Cat~II) is of lesser importance. It is just an intermediate step before the $\mathrm{3^{rd}}$ and most specific level which gathers identical products or objects. Examples of those $\mathrm{3^{rd}}$ level categories~(Cat~III) belonging to the 3 stores mentioned above would be BABY BOTTLE (BIBERON), \textit{TRAVEL BAG} (SAC DE VOYAGE) and \textit{PHOTO FRAME} \textit{(CADRE PHOTO)}. The number of categories for each level is given in table~\ref{table:cat}. A ratio of roughly 1 to 10 is observed in the number of categories from one level to the next, leading to 5270~distinct Cat~III categories. It is worth noting that there are actually 5263~distinct values taken by those 5270 Cat~III categories: 7~couples of them share the same name while belonging to different Cat~II categories. However, the combination Cat~II \& Cat~III is uniquely defined through the dataset. Finally, each of the 5270 Cat~I \& Cat~II \& Cat~III category is encoded with an integer index in the dataset. 

Down to the level of products, we count between 1~and 4~180x180~images that can be associated to a given product. There aren't any specific rule to define that number as within our Market Place, a reseller is simply given the choice to insert 1, 2, 3 or 4~images with his products. The table~\ref{table:img} summarizes the distribution of products according to the number of associated images. More than half the number of products have only 1~image. But this reduces to $1/3$~the number of images in the dataset. Finally, we count precisely 12,371,293~images for 7,069,896~products.

\begin{figure*}
	\centering
    \includegraphics[width=0.18\textwidth]{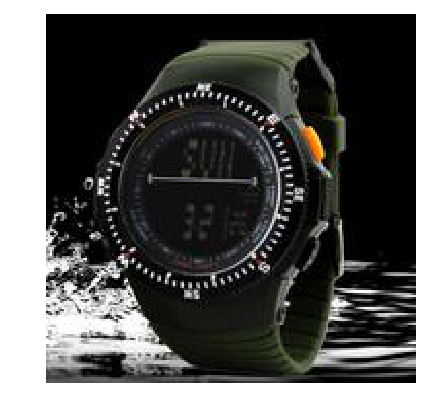}\hspace*{-0.9em} 
    \includegraphics[width=0.18\textwidth]{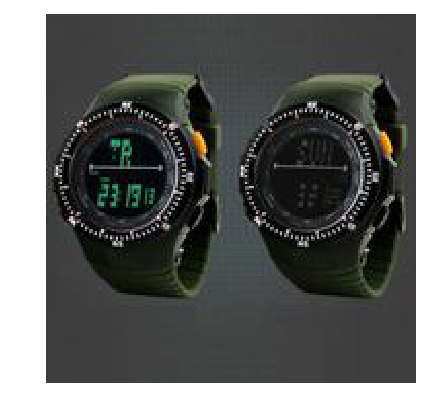}\hspace*{-0.9em}     
    \includegraphics[width=0.18\textwidth]{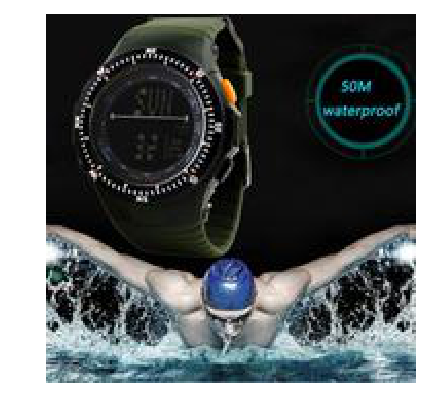}\hspace*{-0.9em}    
    \includegraphics[width=0.18\textwidth]{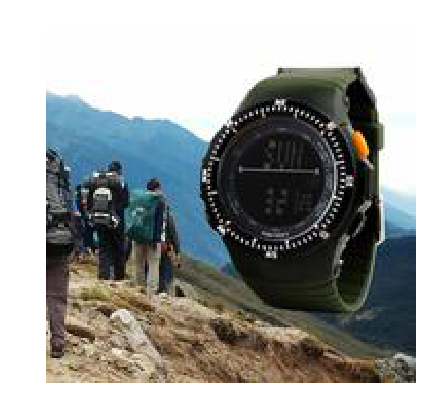}\hspace*{-0.9em}     
    \includegraphics[width=0.18\textwidth]{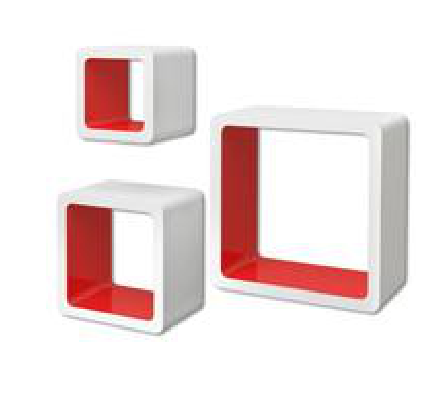}\hspace*{-0.9em}     
    \includegraphics[width=0.18\textwidth]{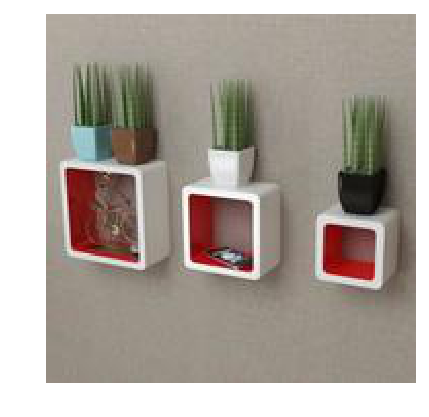}    

	\centering
    \includegraphics[width=0.18\textwidth]{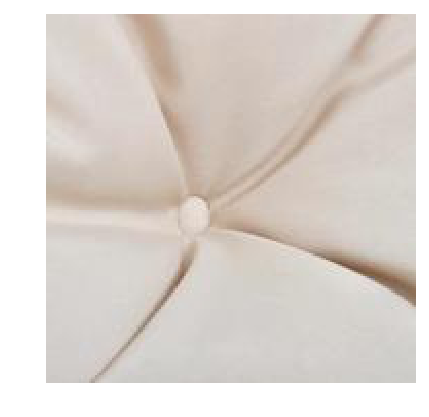}\hspace*{-0.9em}     
    \includegraphics[width=0.18\textwidth]{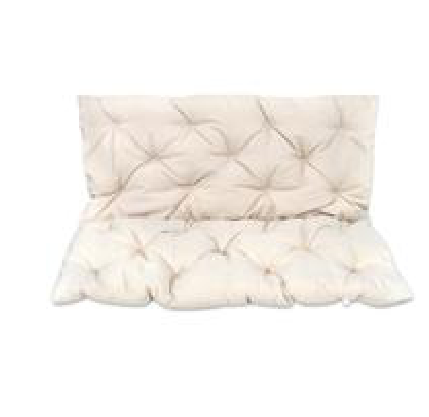}\hspace*{-0.9em}    
    \includegraphics[width=0.18\textwidth]{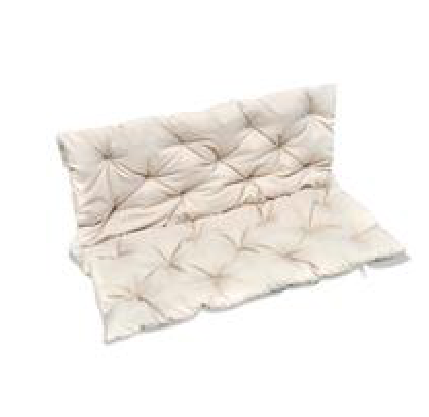}\hspace*{-0.9em}     
    \includegraphics[width=0.18\textwidth]{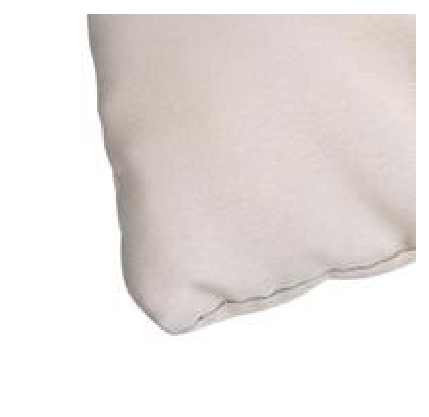}\hspace*{-0.9em}     
    \includegraphics[width=0.18\textwidth]{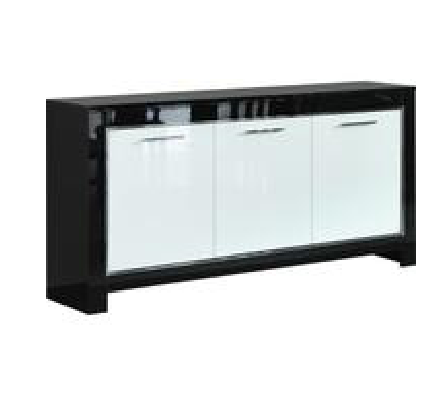}\hspace*{-0.9em}     
    \includegraphics[width=0.18\textwidth]{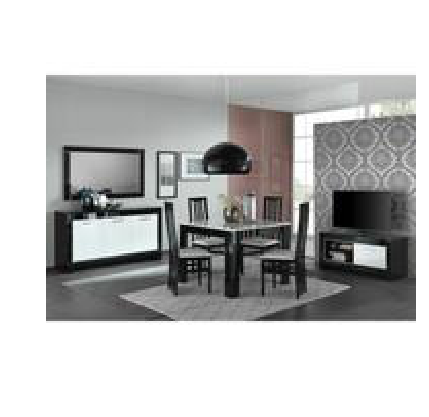}    
  
	\centering
    \includegraphics[width=0.18\textwidth]{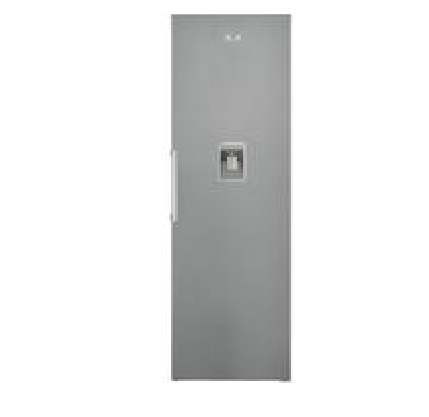}\hspace*{-0.9em}    
    \includegraphics[width=0.18\textwidth]{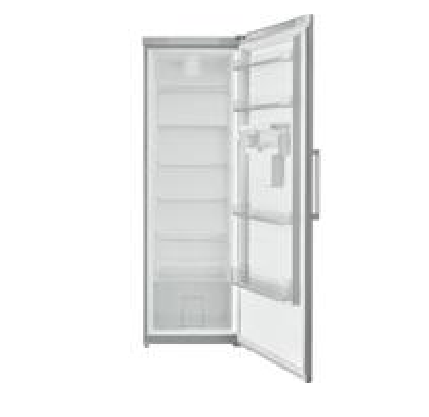}\hspace*{-0.9em}    
    \includegraphics[width=0.18\textwidth]{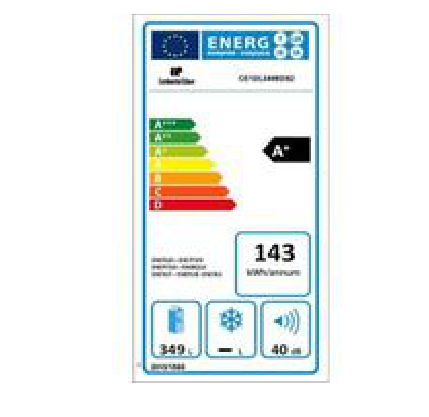}\hspace*{-0.9em}    
    \includegraphics[width=0.18\textwidth]{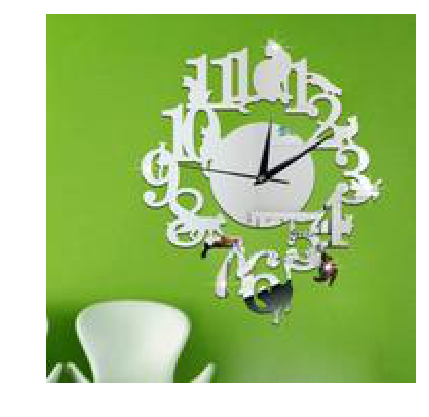}\hspace*{-0.9em}    
    \includegraphics[width=0.18\textwidth]{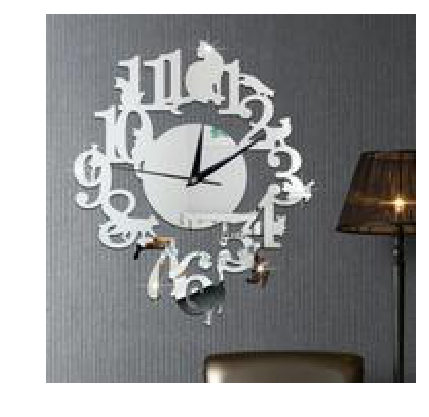}\hspace*{-0.9em}    
    \includegraphics[width=0.18\textwidth]{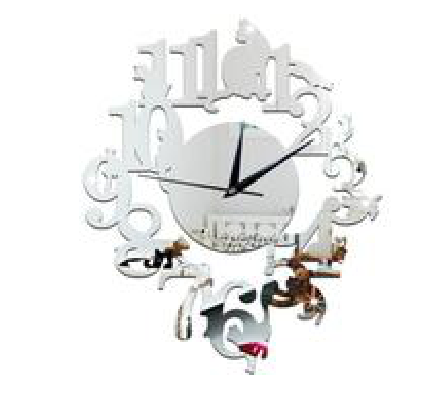}    
  
	\centering
    \includegraphics[width=0.18\textwidth]{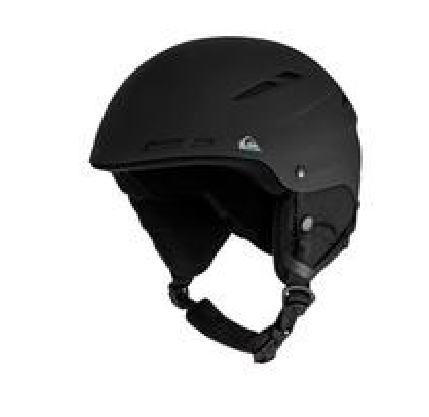}\hspace*{-0.9em}    
    \includegraphics[width=0.18\textwidth]{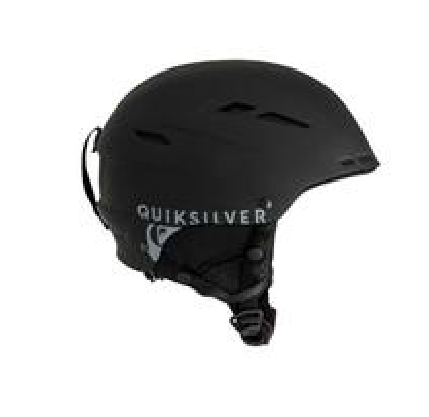}\hspace*{-0.9em}    
    \includegraphics[width=0.18\textwidth]{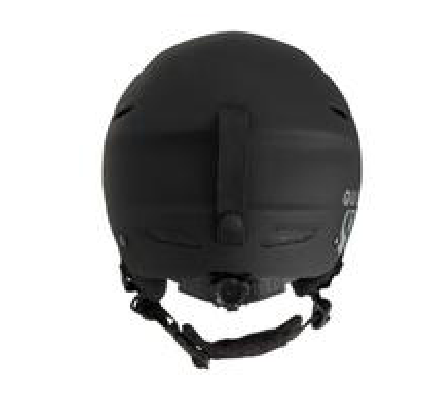}\hspace*{-0.9em}    
    \includegraphics[width=0.18\textwidth]{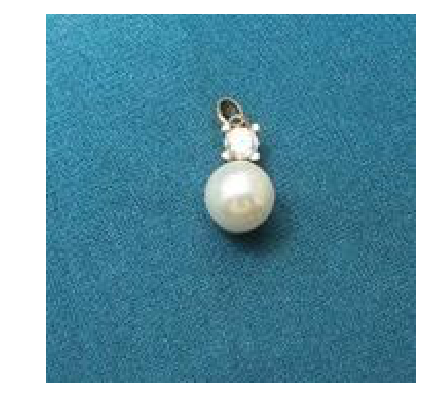}\hspace*{-0.9em}    
    \includegraphics[width=0.18\textwidth]{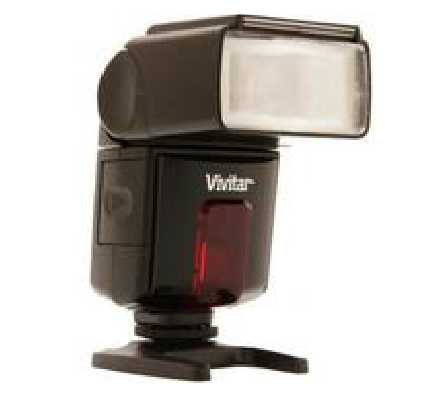}\hspace*{-0.9em}    
    \includegraphics[width=0.18\textwidth]{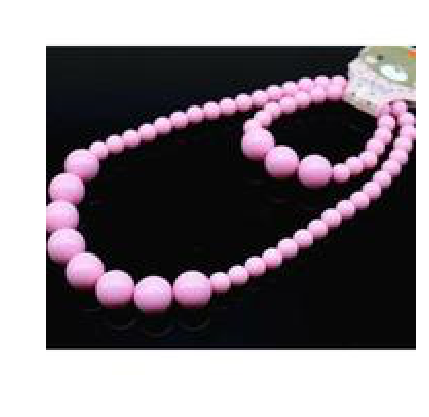}    
  \caption{Examples of images from the dataset. From top to bottom and from left to right: 4 images of a watch, 2 images of a wall decoration; 4 images of a couch, 2 images of a dresser;  3 images associated to a fridge (2 views and 1 label), 3 images of a clock; 3 images of a helmet, 1 image of a neck-lace, 1 image of a camera flash, 1 image of a pendant.
  }
  \label{fig:example}
\end{figure*}

The initial product classification on which was created the dataset was made using textual descriptions we have for each product in our catalog. The process of classification is semi-automatized: a K-NN is applied to classify every product and if the required confidence level isn't met for a given product, it is sent to manual classification. Finally, the overall quality of the classification is assessed by frequent sampling operations in which a trained expert is asked to visually control the classification. The measured overall rate of bad classification based on this sampling technique is around 10~\% in each category. This number gives the order of magnitude of the noise associated to our image dataset.

The figure~\ref{fig:example} shows some illustrative examples of images that can be found in the dataset. The background may vary a lot from one image to the other. A product might be presented on a white or colored background or they might be shown in an illustrative situation like the wall decoration or the dresser. Images might be views of the same object with different angles like for the helmet or they might be showing a zoom on some specific detail of the product as for the couch. The product may also be represented more than once like the watch. Finally, for some specific products, one of the images might not be showing the product at all. This is the case for the fridge as, according to the European Union regulation, electrical goods all have to carry an EU energy label.

\section{Detailed features of the dataset}\label{sec-EDA}

In this section, we focus on 2 characteristics of the dataset that one should consider when using it to train a neural network. The first one is the unbalancedness of the dataset and the second is the presence of duplicates among the images.

\subsection{Distribution of products}

\begin{figure*}
  \centering
    \includegraphics[width=0.33\textwidth]{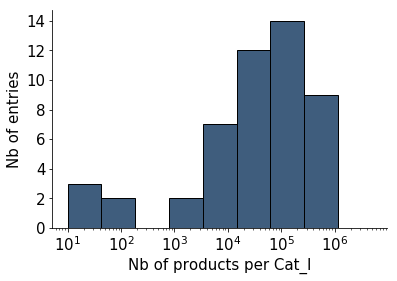}
    \includegraphics[width=0.33\textwidth]{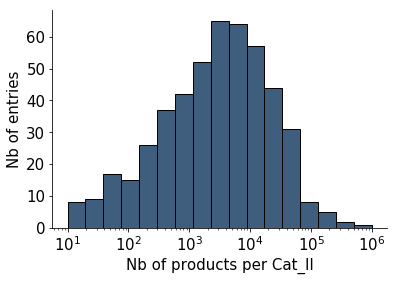}
    \includegraphics[width=0.33\textwidth]{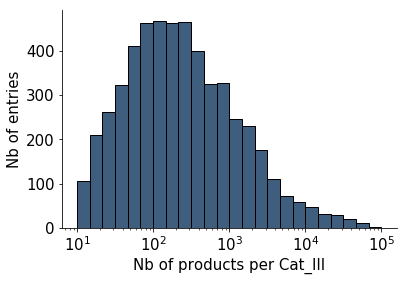}
  \caption{Distributions of products per category for each category level. From left to right: Cat~I, Cat~II and Cat~III.}
  \label{fig:hist_per_cat}
\end{figure*}

Our product catalog is highly diverse and the categorization tree we use is not aimed at balancing the number of products among the categories. Rather, it is aimed at gathering products with similar characteristics and usages. This results in a highly unbalanced number of products per categories.

The figure~\ref{fig:hist_per_cat} shows 3 distributions of products per category, one for each hierarchical level of category. It should be noted that bin widths that were used to draw those histograms vary on a logarithmic scale to facilitate the visualization of several orders of magnitude on the same plot. 

At the Cat~I level, the spread is considerable. There is a small cluster of 5~categories with less than 200~products (APICULTURE, PRODUITS SURGELES, ABONNEMENTS/SERVICES, PRODUITS FRAIS, FUNERAIRE). The rest of the 44~categories gather between roughly $10^4$ and $10^6$ products each. The last bin alone contains 9~categories in which 4.5M products (more than half the products) are to be found in total.

Down to the Cat~II level, the spread remains important. It varies between 10 and half of million of products in just one category named PARTS~(PIECES). The mode of the distribution for this level as shown on figure~\ref{fig:hist_per_cat} is around 3,000~products per category. 

Finally, at the Cat~III level, the most populated categories are rich of more than 10,000~products each (figure~\ref{fig:hist_per_cat}). The top 5 being POP ROCK MUSIC, PRINTER TONER, PRINTER CARTRIDGE, FRENCH LITTERATURE and OTHER BOOKS with nearly 70,000~items each. Most of the categories (nearly 2000~of them) count between 50 and 500~products. 

\begin{figure}
  \centering
    \includegraphics[width=0.45\textwidth]{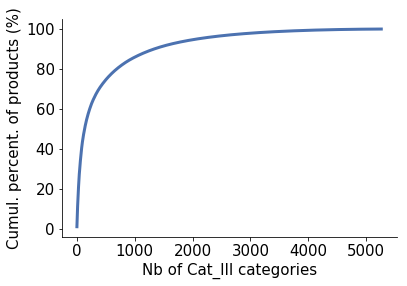}
  \caption{Cumulative percentage of products as a function of the number of Cat~III categories.}
  \label{fig:cumul_per_cat}
\end{figure}

Another way to consider the unbalancedness of the dataset is to consider the share of products with respect to the most populated categories. This is shown in figure~\ref{fig:cumul_per_cat} where the cumulative percentage of products is displayed as a function of the number of Cat~III categories. The behavior is nearly exponential with 75~\% of the products gathered in only 10~\% of the categories. On the other end, the less populated 75~\% of the categories account for only 10~\% of the total number of products. 

\subsection{Duplicated images}

A second aspect specific to this data set might be the presence of duplicated images. Indeed, for a given product there might be several resellers and nothing prevents them from using similar if not identical images to describe their products. From an image classification point of view, the presence of duplicated images might be consider as a downside or an upside. It reduces the absolute size of the dataset but it may also help to classify some products and to make links between categories that contain identical images. 

\begin{figure}
  \centering
    \includegraphics[width=0.45\textwidth]{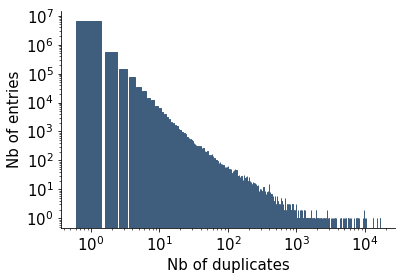}
  \caption{Distribution of images according to their number of duplicates.}
  \label{fig:hist_md5}
\end{figure}

To trace back identical images, we use the MD5 hash function as defined in~\cite{rivest1992md5}. Images with the same hash key are labeled as identical. Although the MD5 hash function suffer weaknesses that will prevent nearly identical images to be detected, it is efficient enough in our case where there is no will to hide duplicates by tricking the image.

The distribution of the measured MD5 hash keys through the entire dataset is shown in figure~\ref{fig:hist_md5}. Again, logarithmic scales have been used to account for the large range of values. Among the 12M~images in total in the dataset, 6.85M~images are uniquely and the 75~\% of the images appear at the most 10~times in the dataset. The image the most replicated appears 16,643~times.

\section{Image classification challenge}\label{sec-challenge}

In this section, we present the image classification challenge that was organized around this dataset and review the 3 winning solutions of the challenge.

\subsection{Description}

The challenge aimed at building an image categorizer based on the dataset presented in this document. It was held between September and December~2017 on the famous Kaggle platform~\cite{kaggle} and involved more than 600~teams from all around the world. The evaluation metric was the accuracy of classification on a test set for which none of the categories is not known (if one wishes to evaluate a categorizer on the test set, please contact the author of the paper). The name, final rank and score of the 3~winning teams is given in the table~\ref{table:LB}. Final results were quite impressive as the best solutions were able to correctly classify almost 80~\% of the images of the test set.

\begin{table}[!htbp]
\center
	\begin{tabular}{l||ccc}
    \textbf{Team name} & bestfitting & convoluted prediction & dylan\\
    \hline
    \textbf{Rank} & $\mathrm{1^{st}}$ & $\mathrm{2^{nd}}$ & $\mathrm{3^{rd}}$\\
    \textbf{Score} & 0.79567 & 0.79352 & 0.79046 \\
    \end{tabular}
    \caption{Name, rank and final score of the 3 winning solutions of the image classification challenge.}
    \label{table:LB}
\end{table}

\subsection{Common features of the 3 winning solutions}

The 3 winning solutions share several characteristics. First, they are all ensemble models that aggregate sub-model in different ways: either they use an elaborate method such as Xgboost~\cite{chen2016xgboost} aggregation method or they simply compute the geometric mean or a weighted mean.

All 3 solutions also rely on neural network architectures pre-trained on ImageNet and fine-tuned on our dataset. Different framework were used : Pytorch~\cite{paszke2017pytorch}, Keras~\cite{chollet2015keras}, Tensorflow~\cite{abadi2016tensorflow} and Mxnet~\cite{chen2015mxnet} and heavy duty GPUs were required to perform transfer learning~\cite{donahue2014decaf}. 

Also, all the state of the art neural network architecture are present :
\begin{itemize}
\item the $\mathrm{1^{st}}$ solution uses ResNet~\cite{he2016deep} and InceptionResNetV2~\cite{szegedy2017inception};
\item the $\mathrm{2^{nd}}$ solution makes a heavy use of squeeze and excitation blocks~\cite{hu2017squeeze} on InceptionV3~\cite{szegedy2017inception}, ResNet~\cite{he2016deep} and ResNExt~\cite{xie2017aggregated} architectures. Dual Path Networks are also used~\cite{chen2017dual};
\item the $\mathrm{3^{rd}}$ solution uses SE blocks on ResNExt architecture and densily connected networks~\cite{huang2017densely}.
\end{itemize}

It is worth noting that none of the above mentioned neural network techniques or architecture were described in the literature before 2016 for a challenge that took place in 2017.

Finally, all 3 solutions made use of the dropout technique~\cite{srivastava2014dropout} to prevent over-fitting and used various technique of data augmentation such as cropping, flipping	and resizing. 

\subsection{Solution specific aspects}

At a more detailed level, the winning solution present an interesting approach in which the trained neural networks are specialized by number of images available for each product. When more than one images are available, they are simply aggregated side by side in a larger image and fed into the dedicated model.

The $\mathrm{2^{nd}}$ solution took a similar approach for one of its sub-model by specializing neural networks for each number and rank of images. Indeed, the competitor realized that both the number of images per product and the order of those images were not random and thus should be considered for classification.

Finally, a last worth mentioning originality was the use by the winning team of an OCR for products that were recognized as being books or CD. This last trick probably made the small difference that made this solution the winning one.  

\section{Conclusion}

A large e-retailer image dataset was presented in this document. This large dataset is publicly released in order to stimulate and leverage the use of deep neural network architectures for creative use cases in recommendation systems. The dataset is made up of a large diversity of product images organized according to a 3 level aggregation tree. We count 12M~images of 7M~products arranged in 5270~categories. It is unbalanced and there are duplicated images in it. An image classification challenge organized around this dataset on the Kaggle platform lead to a classification accuracy of nearly 80~\%.

\bibliography{biblio}

\begin{thebibliography}{25}
\providecommand{\natexlab}[1]{#1}
\providecommand{\url}[1]{\texttt{#1}}
\expandafter\ifx\csname urlstyle\endcsname\relax
  \providecommand{\doi}[1]{doi: #1}\else
  \providecommand{\doi}{doi: \begingroup \urlstyle{rm}\Url}\fi

\bibitem[kag()]{kaggle}
Kaggle platform.
\newblock
  \url{https://www.kaggle.com/c/cdiscount-image-classification-challenge}.

\bibitem[Abadi et~al.(2016)Abadi, Barham, Chen, Chen, Davis, Dean, Devin,
  Ghemawat, Irving, Isard, et~al.]{abadi2016tensorflow}
Mart{\'\i}n Abadi, Paul Barham, Jianmin Chen, Zhifeng Chen, Andy Davis, Jeffrey
  Dean, Matthieu Devin, Sanjay Ghemawat, Geoffrey Irving, Michael Isard, et~al.
\newblock Tensorflow: A system for large-scale machine learning.
\newblock In \emph{OSDI}, volume~16, pages 265--283, 2016.

\bibitem[Chen and Guestrin(2016)]{chen2016xgboost}
Tianqi Chen and Carlos Guestrin.
\newblock Xgboost: A scalable tree boosting system.
\newblock In \emph{Proceedings of the 22nd acm sigkdd international conference
  on knowledge discovery and data mining}, pages 785--794. ACM, 2016.

\bibitem[Chen et~al.(2015)Chen, Li, Li, Lin, Wang, Wang, Xiao, Xu, Zhang, and
  Zhang]{chen2015mxnet}
Tianqi Chen, Mu~Li, Yutian Li, Min Lin, Naiyan Wang, Minjie Wang, Tianjun Xiao,
  Bing Xu, Chiyuan Zhang, and Zheng Zhang.
\newblock Mxnet: A flexible and efficient machine learning library for
  heterogeneous distributed systems.
\newblock \emph{arXiv preprint arXiv:1512.01274}, 2015.

\bibitem[Chen et~al.(2017)Chen, Li, Xiao, Jin, Yan, and Feng]{chen2017dual}
Yunpeng Chen, Jianan Li, Huaxin Xiao, Xiaojie Jin, Shuicheng Yan, and Jiashi
  Feng.
\newblock Dual path networks.
\newblock In \emph{Advances in Neural Information Processing Systems}, pages
  4470--4478, 2017.

\bibitem[Chollet et~al.(2015)]{chollet2015keras}
Fran{\c{c}}ois Chollet et~al.
\newblock Keras, 2015.

\bibitem[Donahue et~al.(2014)Donahue, Jia, Vinyals, Hoffman, Zhang, Tzeng, and
  Darrell]{donahue2014decaf}
Jeff Donahue, Yangqing Jia, Oriol Vinyals, Judy Hoffman, Ning Zhang, Eric
  Tzeng, and Trevor Darrell.
\newblock Decaf: A deep convolutional activation feature for generic visual
  recognition.
\newblock In \emph{International conference on machine learning}, pages
  647--655, 2014.

\bibitem[He et~al.(2016)He, Zhang, Ren, and Sun]{he2016deep}
Kaiming He, Xiangyu Zhang, Shaoqing Ren, and Jian Sun.
\newblock Deep residual learning for image recognition.
\newblock In \emph{Proceedings of the IEEE conference on computer vision and
  pattern recognition}, pages 770--778, 2016.

\bibitem[Hu et~al.(2017)Hu, Shen, and Sun]{hu2017squeeze}
Jie Hu, Li~Shen, and Gang Sun.
\newblock Squeeze-and-excitation networks.
\newblock \emph{arXiv preprint arXiv:1709.01507}, 2017.

\bibitem[Huang et~al.(2017)Huang, Liu, Weinberger, and van~der
  Maaten]{huang2017densely}
Gao Huang, Zhuang Liu, Kilian~Q Weinberger, and Laurens van~der Maaten.
\newblock Densely connected convolutional networks.
\newblock In \emph{Proceedings of the IEEE conference on computer vision and
  pattern recognition}, volume~1, page~3, 2017.

\bibitem[Huang et~al.(2015)Huang, Yong, Li, Yamakawa, and
  Dos~Santos]{huang2015systems}
Joseph Jyh-huei Huang, Chang Yong, Hsiang-Tsun Li, Devender~Akira Yamakawa, and
  Jose~Ricardo Dos~Santos.
\newblock Systems and methods for image recognition using mobile devices,
  November~24 2015.
\newblock US Patent 9,195,898.

\bibitem[Jia et~al.(2009)Jia, Wei, Socher, Li-Jia, Kai, and Li]{Jia09}
D.~Jia, D.~Wei, R.~Socher, L.~Li-Jia, L.~Kai, and F.-F. Li.
\newblock {ImageNet: A large-scale hierarchical image database}.
\newblock In \emph{IEEE Computer Society Conference on Computer Vision and
  Pattern Recognition}, CVPR '09, pages 248--255. IEEE, 2009.
\newblock \doi{10.1109/CVPRW.2009.5206848}.

\bibitem[Kiapour et~al.(2015)Kiapour, Han, Lazebnik, Berg, and Berg]{Kiapour15}
M.H. Kiapour, X.~Han, S.~Lazebnik, A.C. Berg, and T.L. Berg.
\newblock {Where to buy it: matching street clothing photos in online shops}.
\newblock In \emph{2015 IEEE International Conference on Computer Vision},
  ICCV, pages 3343--3351, 2015.
\newblock \doi{10.1109/ICCV.2015.382}.

\bibitem[Krasin et~al.(2017)Krasin, Duerig, Alldrin, Ferrari, Abu-El-Haija,
  Kuznetsova, Rom, Uijlings, Popov, Veit, Belongie, Gomes, Gupta, Sun, Chechik,
  Cai, Feng, Narayanan, and Murphy]{Krasin17}
I.~Krasin, T.~Duerig, N.~Alldrin, V.~Ferrari, S.~Abu-El-Haija, A.~Kuznetsova,
  H.~Rom, J.~Uijlings, S.~Popov, A.~Veit, S.~Belongie, V.~Gomes, A.~Gupta,
  C.~Sun, G.~Chechik, D.~Cai, Z.~Feng, D.~Narayanan, and K.~Murphy.
\newblock {OpenImages: A public dataset for large-scale multi-label and
  multi-class image classification.}
\newblock \emph{Dataset available from https://github.com/openimages}, 2017.

\bibitem[Lin et~al.(2014)Lin, Maire, Belongie, Hays, Perona, Ramanan,
  Doll{\'{a}}r, and {Lawrence Zitnick}]{Lin14}
T.Y. Lin, M.~Maire, S.~Belongie, J.~Hays, P.~Perona, D.~Ramanan,
  P.~Doll{\'{a}}r, and C.~{Lawrence Zitnick}.
\newblock {Microsoft COCO: Common objects in context}.
\newblock In D.~Fleet, T.~Pajdla, B.~Schiele, and T.~Tuytelaars, editors,
  \emph{Computer Vision – ECCV 2014}, volume 8693 of \emph{ECCV 2014}, pages
  740--755. Springer, Cham, 2014.
\newblock \doi{10.1007/978-3-319-10602-1_48}.

\bibitem[Liu et~al.(2016)Liu, Luo, Qiu, Wang, and Tang]{Liu16}
Z.~Liu, P.~Luo, S.~Qiu, X.~Wang, and X.~Tang.
\newblock {DeepFashion: powering robust clothes recognition and retrieval with
  rich annotations}.
\newblock In \emph{2016 IEEE Conference on Computer Vision and Pattern
  Recognition}, CVPR 2016, pages 1096--1104, 2016.
\newblock \doi{10.1109/CVPR.2016.124}.

\bibitem[Paszke et~al.(2017)Paszke, Gross, Chintala, and
  Chanan]{paszke2017pytorch}
Adam Paszke, Sam Gross, Soumith Chintala, and Gregory Chanan.
\newblock Pytorch, 2017.

\bibitem[Rivest(1992)]{rivest1992md5}
Ronald Rivest.
\newblock The md5 message-digest algorithm.
\newblock 1992.

\bibitem[Russell et~al.(2008)Russell, Torralba, Murphy, and Freeman]{Russell08}
B.C. Russell, A.~Torralba, Ke.P. Murphy, and W.T. Freeman.
\newblock {LabelMe: a database and web-based tool for image annotation}.
\newblock \emph{International Journal of Computer Vision}, 77\penalty0
  (1-3):\penalty0 157--173, 2008.
\newblock ISSN 09205691.
\newblock \doi{10.1007/s11263-007-0090-8}.

\bibitem[Simonyan and Zisserman(2015)]{simonyan2015very}
Karen Simonyan and Andrew Zisserman.
\newblock Very deep convolutional networks for large-scale image recognition.
  arxiv prepr arxiv14091556.
\newblock 2015.

\bibitem[Srivastava et~al.(2014)Srivastava, Hinton, Krizhevsky, Sutskever, and
  Salakhutdinov]{srivastava2014dropout}
Nitish Srivastava, Geoffrey Hinton, Alex Krizhevsky, Ilya Sutskever, and Ruslan
  Salakhutdinov.
\newblock Dropout: A simple way to prevent neural networks from overfitting.
\newblock \emph{The Journal of Machine Learning Research}, 15\penalty0
  (1):\penalty0 1929--1958, 2014.

\bibitem[Szegedy et~al.(2017)Szegedy, Ioffe, Vanhoucke, and
  Alemi]{szegedy2017inception}
Christian Szegedy, Sergey Ioffe, Vincent Vanhoucke, and Alexander~A Alemi.
\newblock Inception-v4, inception-resnet and the impact of residual connections
  on learning.
\newblock In \emph{AAAI}, volume~4, page~12, 2017.

\bibitem[Torralba et~al.(2008)Torralba, Fergus, and Freeman]{Torralba08}
A.~Torralba, R.~Fergus, and W.T. Freeman.
\newblock 80 million tiny images: a large dataset for non-parametric object and
  scene recognition.
\newblock \emph{IEEE Transactions on Pattern Analysis and Machine
  Intelligence}, 30\penalty0 (11):\penalty0 1958--1970, 2008.
\newblock \doi{10.1109/TPAMI.2008.128}.

\bibitem[Xie et~al.(2017)Xie, Girshick, Doll{\'a}r, Tu, and
  He]{xie2017aggregated}
Saining Xie, Ross Girshick, Piotr Doll{\'a}r, Zhuowen Tu, and Kaiming He.
\newblock Aggregated residual transformations for deep neural networks.
\newblock In \emph{Computer Vision and Pattern Recognition (CVPR), 2017 IEEE
  Conference on}, pages 5987--5995. IEEE, 2017.

\bibitem[Yao et~al.(2007)Yao, Yang, and Zhu]{Yao07}
B.~Yao, X.~Yang, and S.-C. Zhu.
\newblock {Introduction to a large-scale general purpose ground truth database:
  methodology, annotation tool and benchmarks}.
\newblock In \emph{International Workshop on Energy Minimization Methods in
  Computer Vision and Pattern Recognition}, volume 4679 of \emph{EMMCVPR 2007},
  pages 169 -- 183, 2007.
\newblock \doi{10.1007/978-3-540-74198-5}.

\end{thebibliography}

\end{document}